\documentclass[10pt,conference]{IEEEtran}

\RequirePackage{snapshot}

\usepackage{caption}
\usepackage{subcaption}
\usepackage{fancyhdr}
\usepackage{authblk}

\usepackage{moreverb}
\usepackage{verbatim}

\usepackage{acronym}
\usepackage{enumitem}
\usepackage{color}

\usepackage{parskip}

\usepackage{latexsym}
\usepackage{tabularx}
\usepackage{cite} 
\usepackage[sharp]{easylist}
\usepackage{etoolbox}

\usepackage{siunitx}
\usepackage[usenames,dvipsnames,table]{xcolor}

\usepackage{amsmath}
\usepackage{amsfonts}
\usepackage{amssymb}
\usepackage{amsthm}

\usepackage{algorithmic}
\usepackage{algorithm}
\usepackage{cases}

\usepackage[colorinlistoftodos,disable]{todonotes}

\usepackage{styles/mathdefs}

\renewcommand{\eqref}[1]{(\ref{eq:#1})}
\newcommand{\secref}[1]{\S\ref{sec:#1}}

\newcommand{\figref}[1]{Fig.~\ref{fig:#1}}
\newcommand{\tabref}[1]{Table~\ref{tab:#1}}

\newcommand{\rc}[1]{{\color{black}#1}}

\let\OldEasylist\easylist
\let\OldEndEasylist\endeasylist
\renewenvironment{easylist}{%
    \OldEasylist%
    \ListProperties(Style**=\color{blue},Start1=1)%
}{%
    \OldEndEasylist%
}%

\fancypagestyle{firststyle}{
  \fancyhf{} 
  
   \fancyfoot[L]{\textit{This manuscript was submitted to IET Computer Vision on July 30, 2015}}
}

\begin{document}

\title{Owl and Lizard: Patterns of Head Pose and Eye Pose in Driver Gaze Classification\\
}




\author[1]{Lex Fridman}
\author[1]{Joonbum Lee}
\author[1]{Bryan Reimer}
\author[2]{Trent Victor}
\affil[1]{Massachusetts Institute of Technology (MIT)}
\affil[2]{Chalmers University of Technology, SAFER}

\maketitle

\begin{abstract}%

  Accurate, robust, inexpensive gaze tracking in the car can help keep a driver safe by facilitating the more effective
  study of how to improve (1) vehicle interfaces and (2) the design of future Advanced Driver Assistance Systems. In
  this paper, we estimate head pose and eye pose from monocular video using methods developed extensively in prior work
  and ask two new interesting questions. First, how much better can we classify driver gaze using head and eye pose
  versus just using head pose? Second, are there individual-specific gaze strategies that strongly correlate with how
  much gaze classification improves with the addition of eye pose information? We answer these questions by evaluating
  data drawn from an on-road study of 40 drivers. The main insight of the paper is conveyed through the analogy of an
  ``owl'' and ``lizard'' which describes the degree to which the eyes and the head move when shifting gaze. When the
  head moves a lot (``owl''), not much classification improvement is attained by estimating eye pose on top of head
  pose. On the other hand, when the head stays still and only the eyes move (``lizard''), classification accuracy
  increases significantly from adding in eye pose. We characterize how that accuracy varies between people, gaze
  strategies, and gaze regions.
\end{abstract}

\begin{IEEEkeywords}%
Head pose estimation, pupil detection, gaze tracking, driver distraction, driver assistance systems, on-road study.
\end{IEEEkeywords}


\section{Introduction}\label{sec:introduction}

\begin{figure*}
  \centering
  \includegraphics[width=\textwidth]{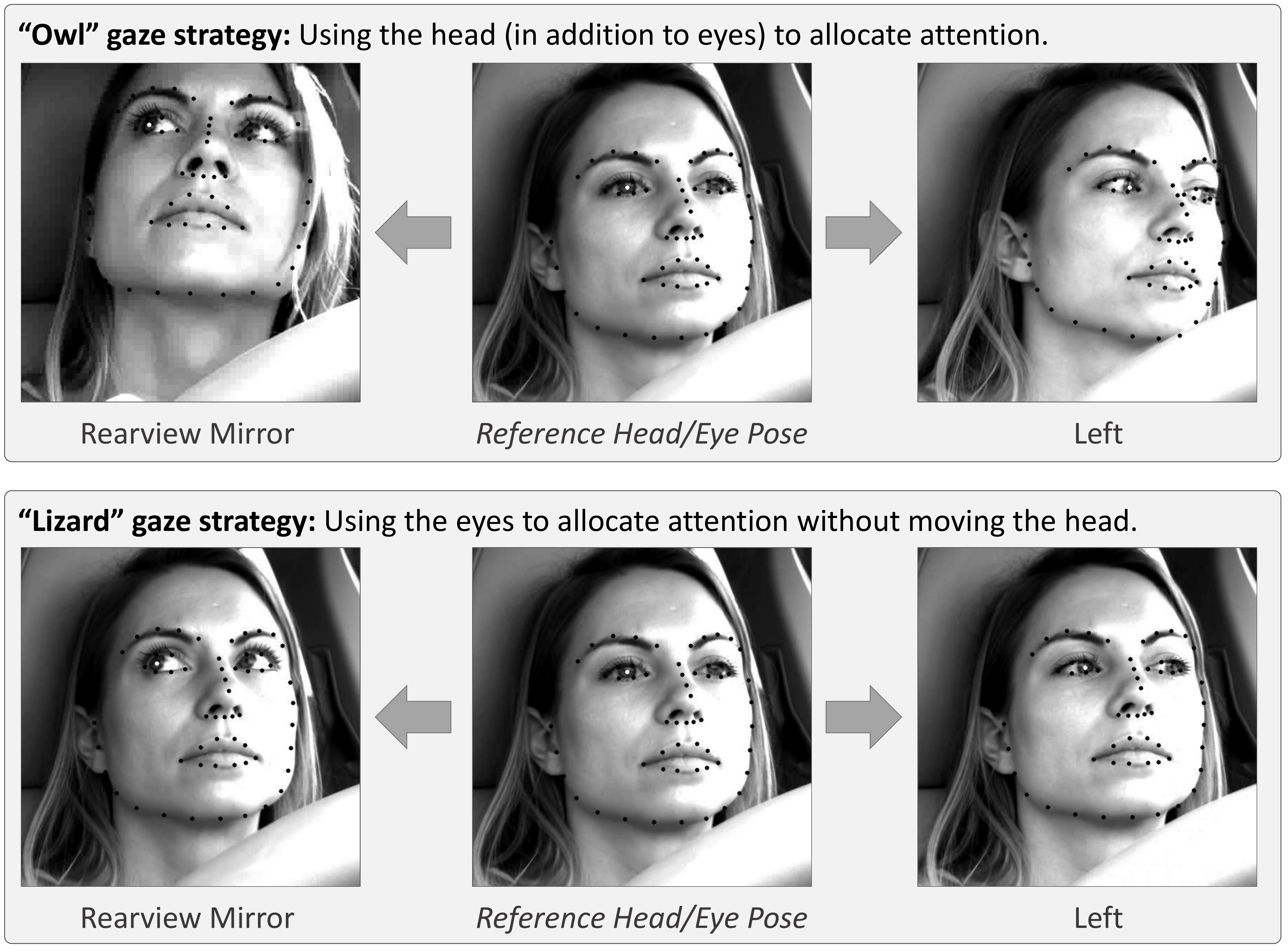}
  \caption{Examples of gaze strategies that explain the ``owl'' and ``lizard'' analogy for head and eye pose. The
    ``owl'' gaze strategy involves primarily head movement, while the ``lizard'' gaze strategy involves primarily eye
    movement. The spectrum between these two is discussed in \secref{results-user-based}.}
  \label{fig:example-owl-lizard}
\end{figure*}

The classification of driver visual attention allocation is an area of increasing relevance in the pursuit of accident
reduction. The allocation of visual attention away from the road has been linked to accident risk
\cite{klauer2006impact,liang2012dangerous} and a drop in situational awareness as uncertainty in the environment
increases \cite{senders1967attentional}. Driver distraction is often construed as a key source of attention divergence
from the roadway and the topic of numerous scientific studies and design guidelines
\cite{national2012visual,driver2006statement}.

Furthermore, as the level of vehicle automation continues to increase through Advanced Driver Assistance Systems as well
as other higher forms of automation, freeing available resources from the primary operational task, drivers are expected
to be increasingly allowed to glance away from the roadway for greater periods. When the need arises to orient the
driver to the roadway, different alerting strategies may be advantageous. Such work would suggest that a real-time
estimation of driver’s gaze could be coupled with an alerting system to enhance safety
\cite{coughlin2011monitoring}. Gaze tracking from video in the driving context is a difficult problem due especially to
rapidly varying lighting conditions. Other challenges, common to other domains, include unpredictability of the
environment, presence of eyeglasses or sunglasses occluding the eye, partial occlusion of the pupil due to squinting,
vehicle vibration, image blur, poor video resolution, etc. We consider the challenging case of uncalibrated monocular
video because it has been and continue to be the most commonly available form of video in driving datasets due to its
low equipment and installation costs.

From the perspective of image processing, gaze estimation can be divided into two components: head pose estimation and
eye pose estimation. Due to all the factors above, the latter is more difficult than the former. In fact, gaze
classification performance can be good based on head pose alone \cite{fridman2015noeye}, because it frequently
correlates with eye pose, but not always. ``Eye pose'' and ``head pose'' are terms used throughout this paper to mean
the relative orientation of the pupil in the eye socket and the relative orientation of facial features on the head,
respectively. This use of ``pose'' is made broader in order to allow for the nonlinear modeling discussed in
\secref{head-pose}.

In this paper, we seek to characterize when eye pose significantly contributes to gaze classification and when it does
not. Specifically, we ask two questions:

\begin{enumerate}
\item \textbf{Contribution of Eye Pose:} How much better can we classify driver gaze using (a) head and eye pose
  together versus (b) using head pose alone.
\item \textbf{Classification of Different Gaze Strategies:} Are there individual-specific gaze strategies that strongly
  correlate with how much gaze classification improves with the addition of eye pose information?
\end{enumerate}

These two questions are answered by analyzing data drawn from an on-road study of 40 drivers performing secondary tasks
of varying complexity. The inter-person classification and gaze strategy variation is discussed using the analogy of an
``owl'' and ``lizard'' (introduced previously in \cite{victor2014analysis,joonbum2015head}) which describes the degree to which the eyes and the
head move when shifting gaze. When the head moves a lot (``owl''), not much classification improvement is attained by
estimating eye pose on top of head pose. On the other hand, when the head stays still and only the eyes move
(``lizard''), classification accuracy increases significantly from adding in eye pose. Examples of the two strategies
are shown in \figref{example-owl-lizard}. We propose an end-to-end driver gaze classification system based on monocular
video and use it to explore the importance of eye pose for classification performance as we move along the spectrum of
people from ``owl'' to ``lizard''.

\section{Related Work}\label{sec:related-work}

The problem of gaze tracking from monocular video has been investigated extensively across many domains
\cite{gaur2014survey, sireesha2013survey}. We build on this work to characterize the individual contribution of head
movement and eye movement to gaze classification accuracy. The building blocks of our image processing pipeline are:
face alignment, head pose estimation, and pupil detection. We apply cutting-edge algorithms from these fields to answer
two questions posed by our work (see \secref{introduction}) on a large on-road driving dataset.

The algorithm in \cite{kazemi2014one} uses an ensemble of regression trees for super-real-time face alignment. Our face
feature extraction algorithm draws upon this method as it is built on a decade of progress on the face alignment problem
(see \cite{kazemi2014one} for a detailed review of prior work). The key contribution of the algorithm is an iterative
transform of the image to a normalized coordinate system based on the current estimate of the face shape. Also, to avoid
the non-convex problem of initially matching a model of the shape to the image data, the assumption is made that the
initial estimate of the shape can be found in a linear subspace.

Head pose estimation has a long history in computer vision. Murphy-Chutorian and Trivedi \cite{murphy2009head} describe
74 published and tested systems from the last two decades. Generally, each approach makes one of several assumptions
that limit the general applicability of the system in driver gaze detection. These assumptions include: (1) the video
is continuous, (2) initial pose of the subject is known, (3) there is a stereo vision system available, (4) the camera
has frontal view of the face, (5) the head can only rotate on one axis, (6) the system only has to work for one
person. While the development of a set of assumptions is often necessary for the classification of a large number of
possible poses, our approach skips the head pose estimation step (i.e. the computation of a vector in 3D space modeling
the orientation of the head) and goes straight from the detection of a facial features to a classification of gaze to
one of six glance regions. Prior work has shown that such a classification set is sufficient for the in-vehicle
environment, even under rapidly shifting lighting conditions \cite{fridman2015noeye}.

Pupil detection approaches have been extensively studied. Methods usually track corneal reflection, distinct pupil shape in
combination with edge-detection, characteristic light intensity of the pupil, or a 3D model of the eye to derive an
estimate of an individual's pupil, iris, or eye position \cite{al2013eye}. Our approach uses an adaptive CDF-based
method \cite{asadifard2010automatic} in conjunction with face alignment that significantly narrows the search space.

Studies of the correlation between head and eye movement have shown inter-person variation in the degree to which the
head serves as a proxy for gaze \cite{fridman2015noeye,joonbum2015head}. For example, a previous work tested drivers'
head movements while looking at the ``road'' and the ``center stack'' and found that: (1) drivers' horizontal range of
head movements varied (from 5 to 20 degrees) across individuals along with (2) their mean differences of horizontal head
angles while looking at the two objects (from 0 to 10 degrees) \cite{joonbum2015head}. This paper makes this variation
more explicit by characterizing classification performance with and without eye pose information.

\section{Dataset}\label{sec:dataset}

Training and evaluation is carried out on a dataset of 40 subjects drawn from a larger driving study of 80
subjects that took place on a local interstate highway (see \cite{mehler2015multi} for detailed experimental
methods). For each subject, the collection of data was carried out in one of two vehicles: 2013 Chevrolet Equinox or
Volvo XC60 (randomly assigned). The drivers performed a number of secondary tasks of varying difficulty including using
the voice interface in the vehicle to enter addresses into the navigation system and using the voice interface as well
as manual controls to select phone numbers from a stored phone list.

Both vehicles were instrumented with an array of sensors for assessing driver behavior. The sensor set included a camera
positioned on the dashboard of each vehicle that was intended to capture the driver's face for annotation of glance
behavior. The cameras were positioned off-axis to the driver and in slightly different locations in the two vehicles
(based upon features of the dashboard, etc.). As each driver positioned the seat (electronic in both vehicles)
differently the relative position of the driver in relation to the camera varied somewhat by subject and across each
driver over time (i.e., drivers move continuously in the seat, etc.). The camera was an Allied Vision Tech Guppy Pro
F-125, capturing grayscale images at a resolution of 800x600 and speed of 30fps. The data was double manually annotated
for driver glances transitions during secondary task periods (at a resolution of sub-200ms) into one of 11 classes (road,
center stack, instrument cluster, rearview mirror, left, right, left blindspot, right blindspot, passenger, uncodable,
and other). As detailed in \cite{mehler2015multi}, any discrepancies between the two annotators were meditated by an
arbitrator. \rc{This method of double annotation and mediation of driver gaze has been shown to produce very accurate
annotations that can be effectively used as ground truth for supervised learning approaches \cite{smith2005methodology}.}

\definecolor{lightHyo}{gray}{0.7}
\newcommand{\headerHyo}[1]{\rule[-1.2em]{0em}{3em}\renewcommand{\arraystretch}{1}\begin{tabular}[c]{@{}l@{}}#1\end{tabular}}
\renewcommand{\arraystretch}{1.5}
\begin{table}[h!]
  \centering
  \begin{tabular}{lll}
    \hline
    \headerHyo{Pruning Steps} &
    \headerHyo{Total Frames\\Remaining} &
    \headerHyo{Fraction of\\Original}\\
    \hline
    0. Total Frames Annotated & 1,351,864 & 100\%\\\arrayrulecolor{lightHyo}\hline
    1. Frames with Faces Detected & 1,073,380 & 79.4\%\\\arrayrulecolor{lightHyo}\hline
    2. Frames with Pupils Detected & 833,049 & 61.6\%\\\arrayrulecolor{black}\hline
  \end{tabular}
  \caption{Dataset statistics for the total number of video frames annotated, the number of frames where faces were
    detected, and the number of frames where pupil were detected. Each of these pruning steps are discussed in \secref{algorithm}.}
  \label{tab:dataset-stats}
\end{table}

In this paper, a broad random subset of data was drawn from the initial experiment and the ``left'' and ``left blind
spot'' classes / ``right'', ``right blind spot'', ``passenger'' classes were collapses respectively in to ``left'' and
``right''. Periods that were labeled ``uncodable'' and ``other'' were excluded. Subject pruning was completed to ensure
that every subject under consideration has sufficient training data for each of the six glance regions (road, center
stack, instrument cluster, rearview mirror, left, and right). The threshold for ``sufficient training'' was that each
subject had at least 120 frames of video (where pupils were detected) for each of the six gaze regions.

As shown in \tabref{dataset-stats}, the resulting dataset contains 1,351,864 images each annotated as belonging to one
of six glance regions. The algorithm described in \secref{algorithm} is used for face detection, face alignment, and
pupil detection. The gaze classification approach requires a face and a pupil to be successfully detected in the
image. The filtering procedure is discussed in detail in \secref{algorithm}. Therefore, in the evaluation we include
only the images where a face and a pupil is detected. As the table shows, on average, a face is detected in 79.4\% of
images. 61.6\% of images pass the full image processing pipeline where both a face and a pupil are detected.

\section{Gaze Classification Pipeline}\label{sec:algorithm}

\begin{figure*}
  \centering
  \includegraphics[width=\textwidth]{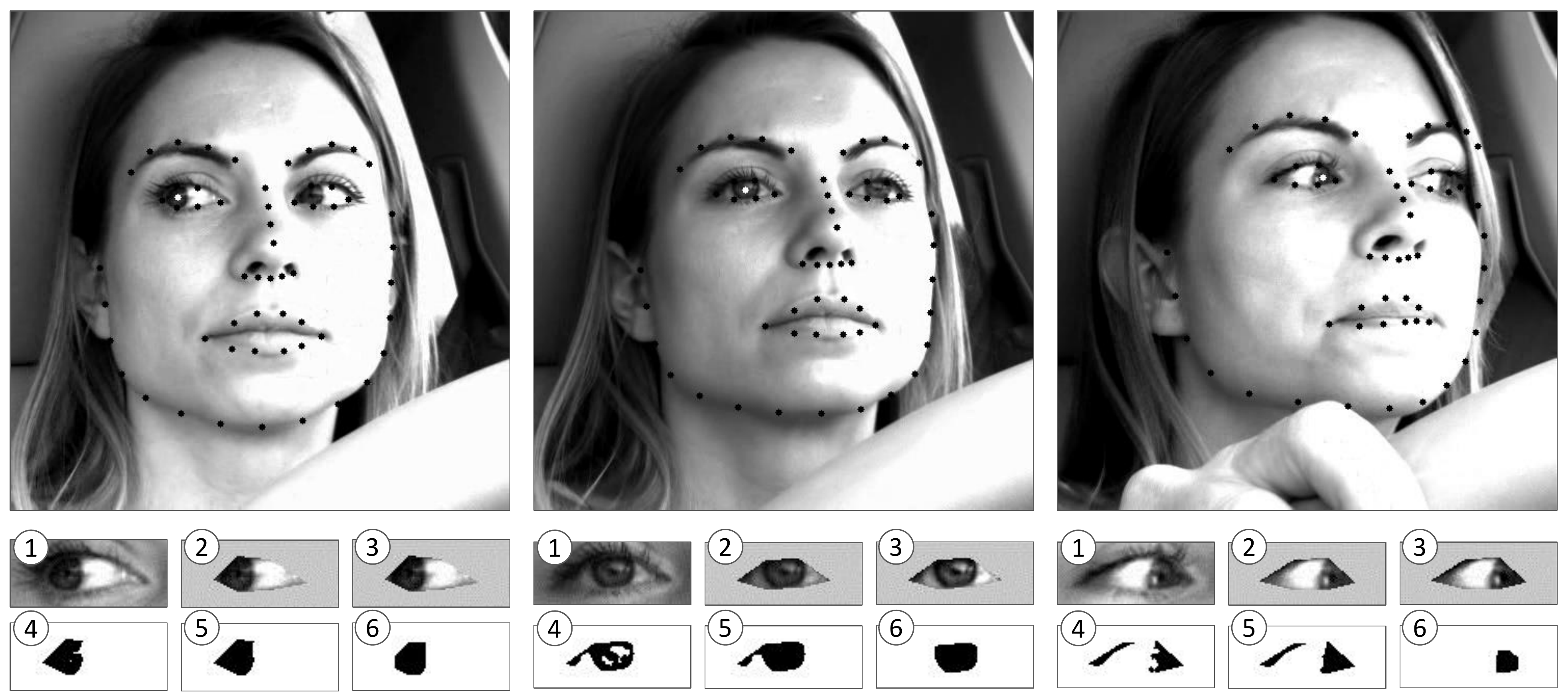}
  \caption{Three example images showing the results of the feature extraction and the intermediate steps of the pupil
    detection. The black dots designate facial landmarks and the single white dot designates the pupil position in the
    right eye. \rc{There are 58 facial landmarks shown with 10 inner-mouth landmarks removed in the visualization for
      the purpose of visual clarity.} Below each of the three face images are 6 steps of the pupil detection described
    in \secref{pupil}.}
  \label{fig:pupil}
\end{figure*}

The steps in the gaze region classification pipeline are: (1) face detection, (2) face alignment, (3) pupil detection,
(4) feature extraction and normalization, (5) classification, (6) decision pruning. If the system passes the first three
steps, it will lead to a gaze region classification decision for every image fed into the pipeline. In step 6, that
decision may be dropped if it falls below a confidence threshold (see \secref{classification}). The three face images in
\figref{pupil} are examples of the result achieved in the first four steps of the pipeline: going from a raw video frame
to extracted face features and pupil position. As mentioned in \secref{introduction}, the relative orientation of facial
features serves as a proxy for ``head pose'' and the relative orientation of pupil position serves as  a proxy for
``eye pose''. We discuss each of the six steps in the pipeline in the following sections.

\subsection{Face Detection}

The environment inside the car is relatively controlled in that the camera position is fixed and the driver torso moves
in a fairly contained space. Thus, a camera can be positioned such that the driver's face is always fully or almost
fully in the field of view. However, the lighting conditions are sometimes drastically variable (e.g. quickly passing
under a bridge, reflection of the sun on the camera lens, etc.) and thus there are frequently cases where the intensity
distribution of the image does not allow for successful detection of the face (i.e. false reject). Every detection step
in the pipeline is tuned to have a low False Accept Rate (FAR). A false accept error early in the pipeline propagates
and can result in drastically incorrect head pose and eye pose estimation. In the context of video-based driver gaze
classification, a high False Reject Rate (FRR) is more acceptable than a high FAR.

The face detector in our pipeline uses a Histogram of Oriented Gradients (HOG) combined with a linear SVM classifier, an
image pyramid, and sliding window detection scheme implemented in the DLIB C++ library \cite{dlib09}. The performance of
this detector has lower FAR than the widely-used default Haar-feature-based face detector available in OpenCV
\cite{lienhart2002extended} and thus is more appropriate for our application.

\subsection{Face Alignment and Head Pose}\label{sec:head-pose}

Both face alignment and head pose estimation are extremely well studied problems in computer vision
\cite{wagner2012toward,murphy2009head}. We investigated several cutting edge methods from each domain, and chose the
ones that worked best for monocular video with highly varying lighting conditions.

Face alignment in our pipeline is performed on a 68-point Multi-PIE facial landmark mark-up used in the iBUG 300-W
dataset \cite{sagonas2013300}. These landmarks include parts of the nose, upper edge of the eyebrows, outer and inner
lips, jawline, and parts in and around the eye. The selected landmarks are shown as black dots in \figref{pupil}. The
algorithm for aligning the 68-point shape to the image data uses a cascade of regressors as described in
\cite{kazemi2014one} and implemented in \cite{dlib09}. The two characteristics of this algorithm most important to
driver gaze localization is: (1) it is robust to partial occlusion and self-occlusion and (2) its running-time is
significantly faster than the 30 fps rate of incoming images.

Face alignment produces estimates for facial feature positions in the image. These features can be mapped directly to a
gaze region using methods that fall under the Nonlinear Regression category defined in \cite{murphy2009head}. They can
also be mapped to a 3d model of the head. The resulting 3D-2D point correspondence can be used to compute the
orientation of the head. This is categorized under Geometric Methods in \cite{murphy2009head}. Then the yaw, pitch, and
roll of the head can be used as features for a gaze region classifier. We implemented both methods and found the former
(nonlinear classification) to be more robust to errors in the face alignment and pupil detection steps of the
pipeline. The geometric approach uses OpenCV's SolvePnP solution of the PnP problem \cite{schweighofer2008globally}. The
nonlinear classification approach is discussed further in \secref{classification}.

\subsection{Pupil Detection}\label{sec:pupil}

As described in \secref{introduction}, the problem of accurate pupil detection is more difficult than the problem of
accurate face alignment, but both are not always robust to poor lighting conditions. Therefore, the secondary task of
pupil detection is to flag errors in the face alignment step that preceded it. As \tabref{dataset-stats} shows, the face
is detected in 79.4\% video frames but only 61.6\% of the original frames pass the pupil detection step.

We use a CDF-based method \cite{asadifard2010automatic} to extract the pupil from the image of the right eye, and adjust
the extracted pupil blob using morphological operations of erosion and dilation. The six steps in this process are as
follows:

\begin{enumerate}
\item Extract the right eye from the face image based on the right eye features computed as part of the face alignment step.
\item Remove all pixels that fall outside the boundaries of the polygon defined by the 6 eye features.
\item Rescale the intensity such that the 98-percentile intensity becomes 1.0 intensity and 2-percentile intensity
  becomes 0.0 intensity.
\item Define a CDF intensity threshold and convert the grayscale image to a binary image. Each pixel intensity above the
  threshold becomes 1, and otherwise becomes 0.
\item Perform an ``opening'' morphology transformation (described in \cite{bradski2008learning}). This operation is
  useful for removing small holes in large blobs.
\item Perform a ``closing'' morphology transformation \rc{(described in \cite{bradski2008learning})}. This operation is useful for removing small objects and
  smoothing the shape of large blobs.
\end{enumerate}

The above steps have three parameters: the CDF threshold, the opening window size, the closing window size. These
parameters are dynamically optimized for each image over a discrete set of values in order to maximize the size of the
largest resulting blobs under one constraint: the largest blob must be circle-shaped (i.e. have similar height and
width).
\rc{More specifically, each of the 3 parameters take on 3 values and using exhaustive search we find the set of
  parameter values that results in the largest circular blob.}

The pupil detection process also includes pruning procedures based on whether the eye is sufficiently open and whether
there is a possible error in the preceding face alignment step. These are:

\begin{enumerate}
\item An eye shape height that is less than 10\% of its width is considered ``closed'' and is removed from the pipeline.
\item When a sufficiently large blob is not found in the eye region \rc{(less than 5 pixels in area)}, it is assumed
  that the face alignment did not properly localize the eye and the image is removed from the pipeline.
\end{enumerate}

\subsection{Feature Extraction and Normalization}

The driver spends more than 90\% of their time looking forward at the road and this fact was used in
\cite{fridman2015noeye} to normalize the position of facial features relative to the average bounding box of the face
associated with the ``Road'' gaze region. This required an initial 120 second period of automated calibration. In this
paper, we remove the need for calibration and instead normalize the facial features based on the bounding box of the
eyes and nose for the current frame only. \rc{\figref{example-owl-lizard} shows departure of the head and eyes away from
  their ``reference'' positions. The normalization step linearly tranforms the facial landmarks such that the landmarks
  of the eyes and nose fit a unit square. After this transformation, the relative orientation of the facial landmarks
  becomes the feature vector for the gaze classification step.}  The bounding box of the eyes and nose was
experimentally found to be the most robust normalizing region. This is due to the fact that the greatest noise in the
face alignment step was associated with the features of the jawline, the eyebrows and the mouth. The position of the
pupil is normalized to the bounding box of the eye rotated such that the two eye corners lie on a horizontal line.

\subsection{Classification and Decision Pruning}\label{sec:classification}

Scikit-learn implementation of a random forest classifier \cite{scikit-learn} is used to generate a set of probabilities
for each class from a single feature vector. The probabilities are computed as the mean predicted class probabilities of
the trees in the forest. The class probability of a single tree is the fraction of samples of the same class in a
leaf. A random forest classifier of depth 25 with an ensemble of 2,000 trees is used for all experiments in
\secref{results}. The class with the highest probability is the one that the system assigns to the image as the
``decision''. The ratio of the highest probability to the second highest probability is termed the ``confidence'' of the
decision. A confidence of 1 is the minimum. There is no maximum. The effect of this threshold is explored in
\cite{fridman2015noeye}. For the experiments in \secref{results} a confidence threshold of 10 is used, which means that
any decisions with a confidence greater than 10 are accepted and the others are ignored. \rc{A random forest classifier
  was used because it achieved a much higher accuracy than k-nearest neigbors (KNN) and linear SVM
  classifiers. RBF-kernel SVM achieved a slightly higher accuracy but at the cost of over a 100-fold increase in
  training time.}

\section{Results}\label{sec:results}

\subsection{Gaze Region Classification}\label{sec:confusion}

\begin{figure}
  \centering
  \includegraphics[width=3.2in]{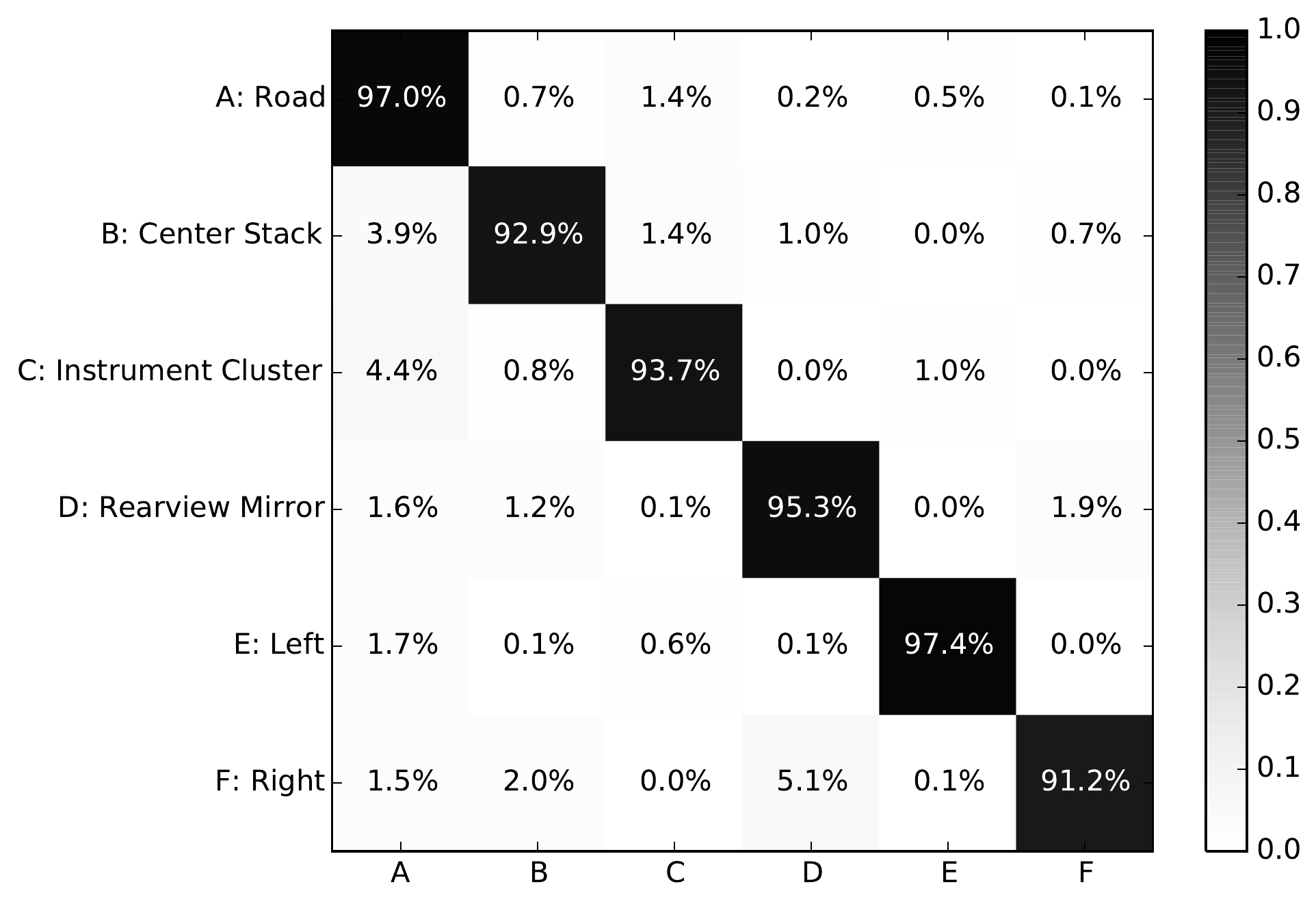}
  \caption{Confusion matrix for the six-region classification problem for using both head pose and eye pose
    information. The overall accuracy is 94.6\%. The confidence threshold is set to 10 resulting in an average confident
    decision rate of 2.3 times a second. When considering that only 61.6\% of frames pass the face detection and pupil
    detection steps, the effective overall decision rate is 1.3 times a second.}
  \label{fig:confusion}
\end{figure}

\begin{figure}
  \centering
  \includegraphics[width=3.2in]{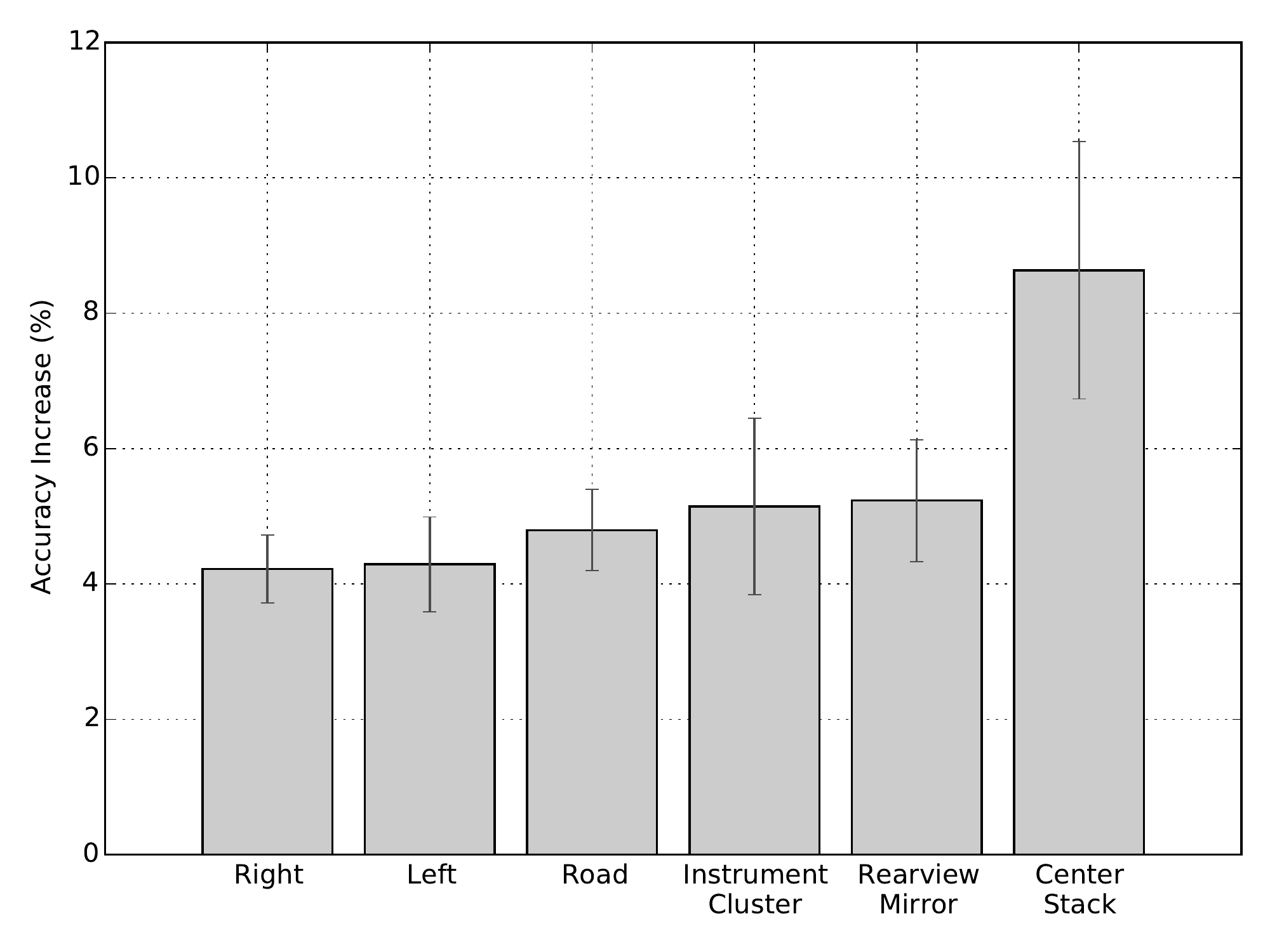}
  \caption{Increase in accuracy per gaze region achieved by using eye pose in addition to head pose. The increase results
  in the confusion matrix show in \figref{confusion}.}
  \label{fig:region-increase}
\end{figure}

We evaluate the gaze classification pipeline described in \secref{algorithm} on the dataset of 40 drivers described in
\secref{dataset}. In all the experiments and discussions that follow, the key comparison is between classification
performed using head pose alone and classification performed using head pose and eye pose together. The classification
problem has six classes, one for each of the six gaze regions: (1) road, (2) center stack, (3) instrument cluster, (4)
rearview mirror, (5) left, and (6) right.

The pipeline starts at an annotated frame from the raw video. As previously described, each frame is double annotated
and mediated ensuring that the gaze region annotations can reliably serve as ground truth for the cross validation
training and testing. There are a total of 1,351,864 annotated images. As show in \tabref{dataset-stats}, 833,049 of
those images pass through the face detection, face alignment, and pupil detection steps of the pipeline. As discussed in
\secref{classification}, we further reduce this number during testing by only considering decisions with a confidence
above the confidence threshold of 10. On average, only 7.1\% of all decisions are deemed confident in this way,
resulting in a decision rate of 2.3 Hz. This selection is distributed evenly through time among cases where a face is
successfully detected. If we consider the fraction of original raw video frames that lead to a confident gaze
classification decision, then the overall effective decision rate is 1.3 Hz.

All of the plots in this sections share the same experiment setup. For each user, we train the 6-class classifier on all
39 others users. The training data for each of the 6 classes is balanced by random sub-sampling
\cite{batista2004study}. The testing is performed on the data for the one user by balancing the classes through
super-sampling. This helps ensure that the per-class accuracy is not skewed by the greater representation of ``Road''
versus the other five classes in the dataset. The process is repeated 100 times for each of the 40 users. The plots with
errorbars indicate the standard deviation of accuracy among the 100 runs for each user.

\figref{confusion} shows the confusion matrix for classification using both head pose and eye pose. The overall accuracy
achieved is 94.6\%. Most of the errors in classification are in incorrectly labeling an image as ``Road'' when it is one
of the other 5 gaze regions. \figref{region-increase} compares the accuracy in this confusion matrix with that achieved
by a system that only uses head pose information. The overall accuracy achieved by such a system is 89.2\%. One of the
questions posed by this paper is: how much to we gain by considering eye pose on top of head pose? The answer in our
final optimized system is 5.4\% accuracy. As \figref{region-increase} shows, the biggest gain of 8.7\% is achieved for
the center stack region. This can be interpreted to mean that people are more likely to use only their eyes when
glancing down to the center stack or that the head pose associated with the center stack is similar to the head pose of
other gaze regions like ``Road'', ``Instrument Cluster'', and ``Rearview Mirror'' as \figref{confusion} suggests.

\subsection{User-Specific Classification and Gaze Strategies}\label{sec:results-user-based}

\begin{figure*}
  \centering
  \begin{subfigure}[t]{0.48\textwidth}
    \includegraphics[height=2.3in]{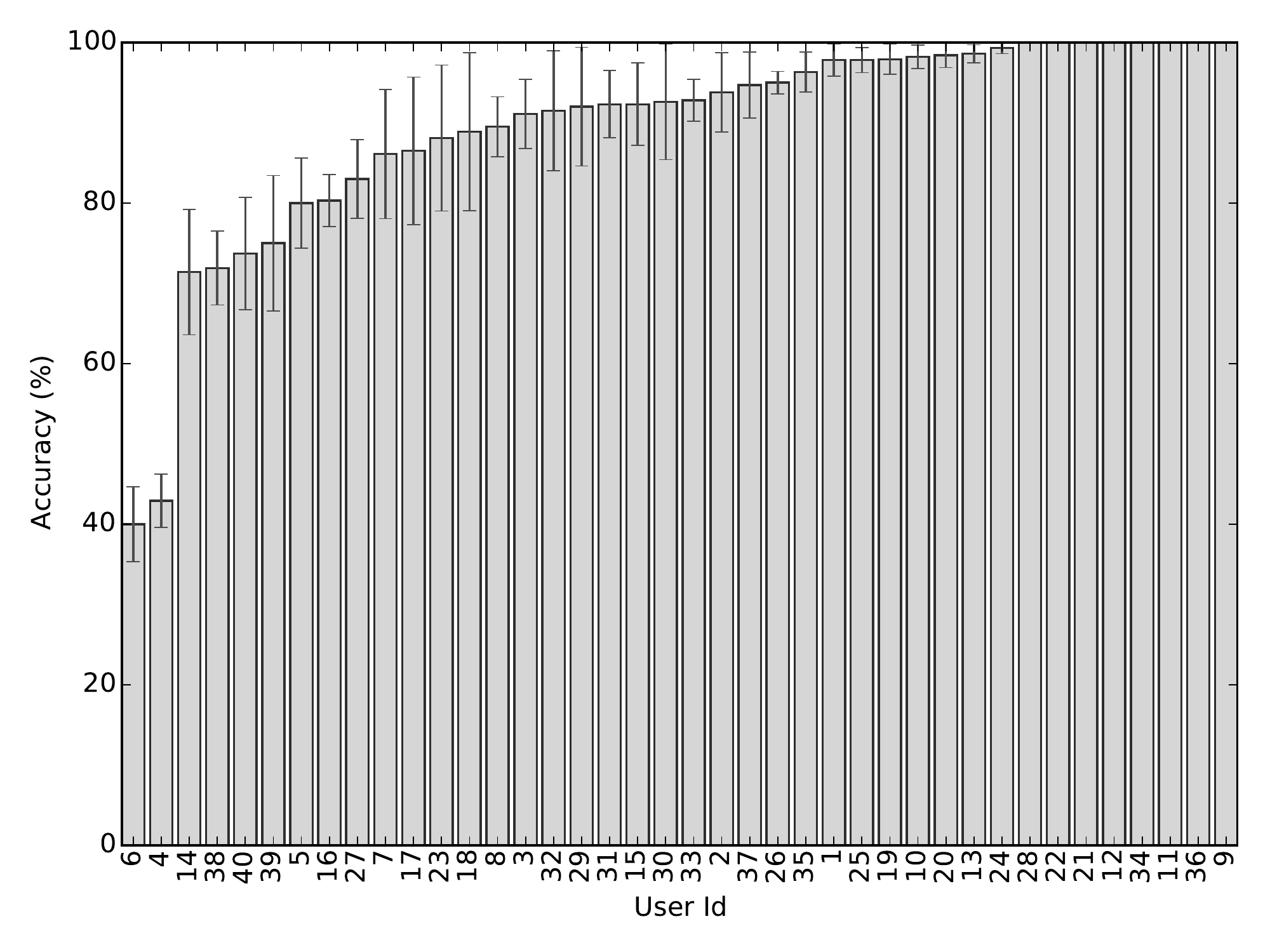}
    \caption{Head pose alone. Average accuracy: 89.2\%}
  \end{subfigure}
  \begin{subfigure}[t]{0.48\textwidth}
    \includegraphics[height=2.3in]{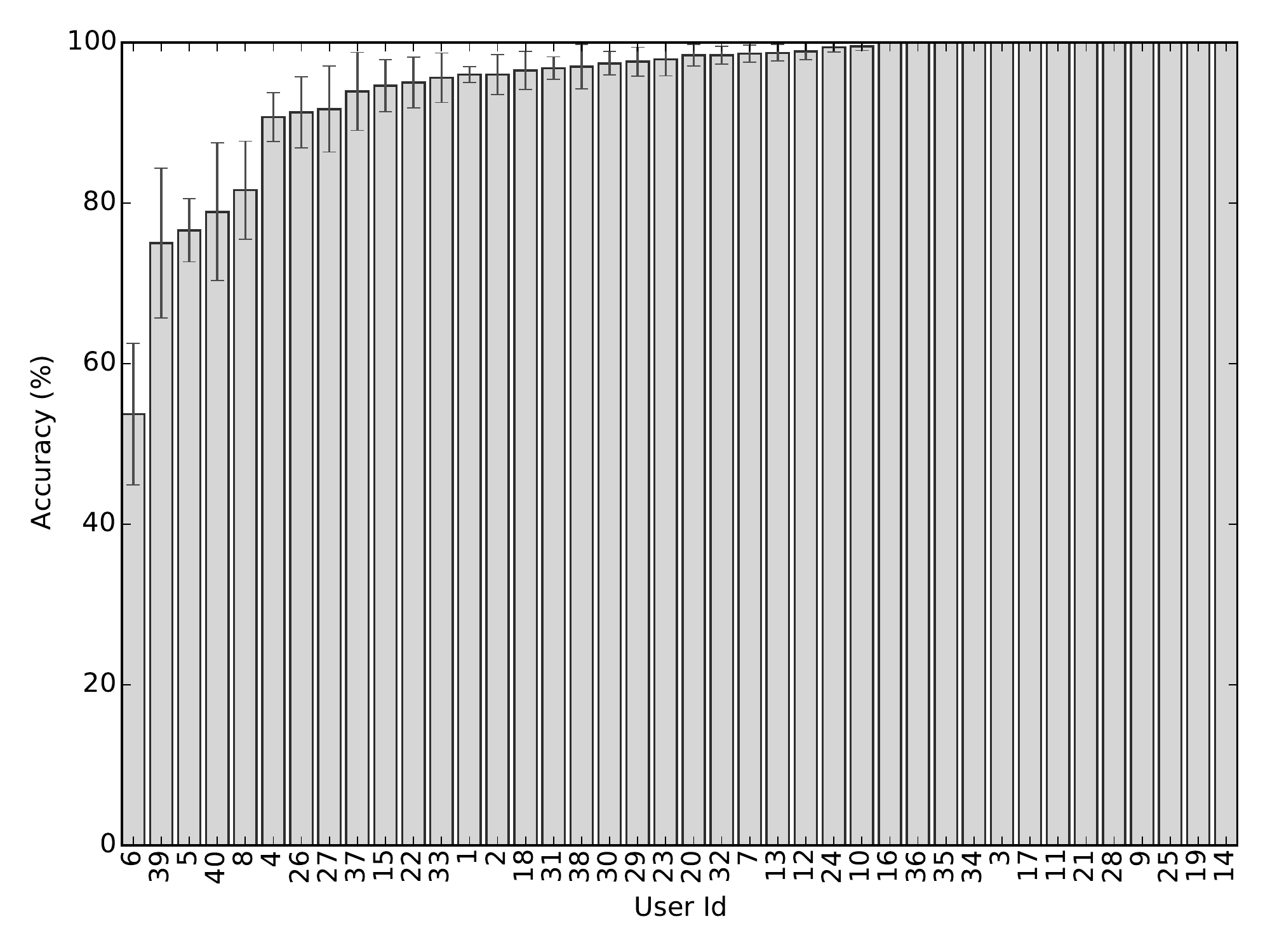}
    \caption{Head pose and eye pose. Average accuracy: 94.6\%}
  \end{subfigure}
  \caption{Per-user accuracy in increasing order for confidence threshold of 10 and resulting decision rate of 2.31
    times a second. The difference between the two is explored further in \figref{per-user-accuracy-increase}.}
  \label{fig:per-user-accuracy}
\end{figure*}

As shown in \secref{confusion}, adding in eye pose to head pose increases gaze classification accuracy by 5.4\%. But
that doesn't tell the full story because some user-trained classifiers benefit more from eye pose than
others. \figref{per-user-accuracy} shows the variation in accuracy among users before and after adding in eye pose to
the classification feature set. \rc{For many users, 100\% accuracy is achieved, while for many other accuracy drops to below
80\% and even to as low as 40\%.}

\begin{figure*}
  \centering
  \begin{subfigure}[t]{0.45\textwidth}
    \includegraphics[height=2.2in]{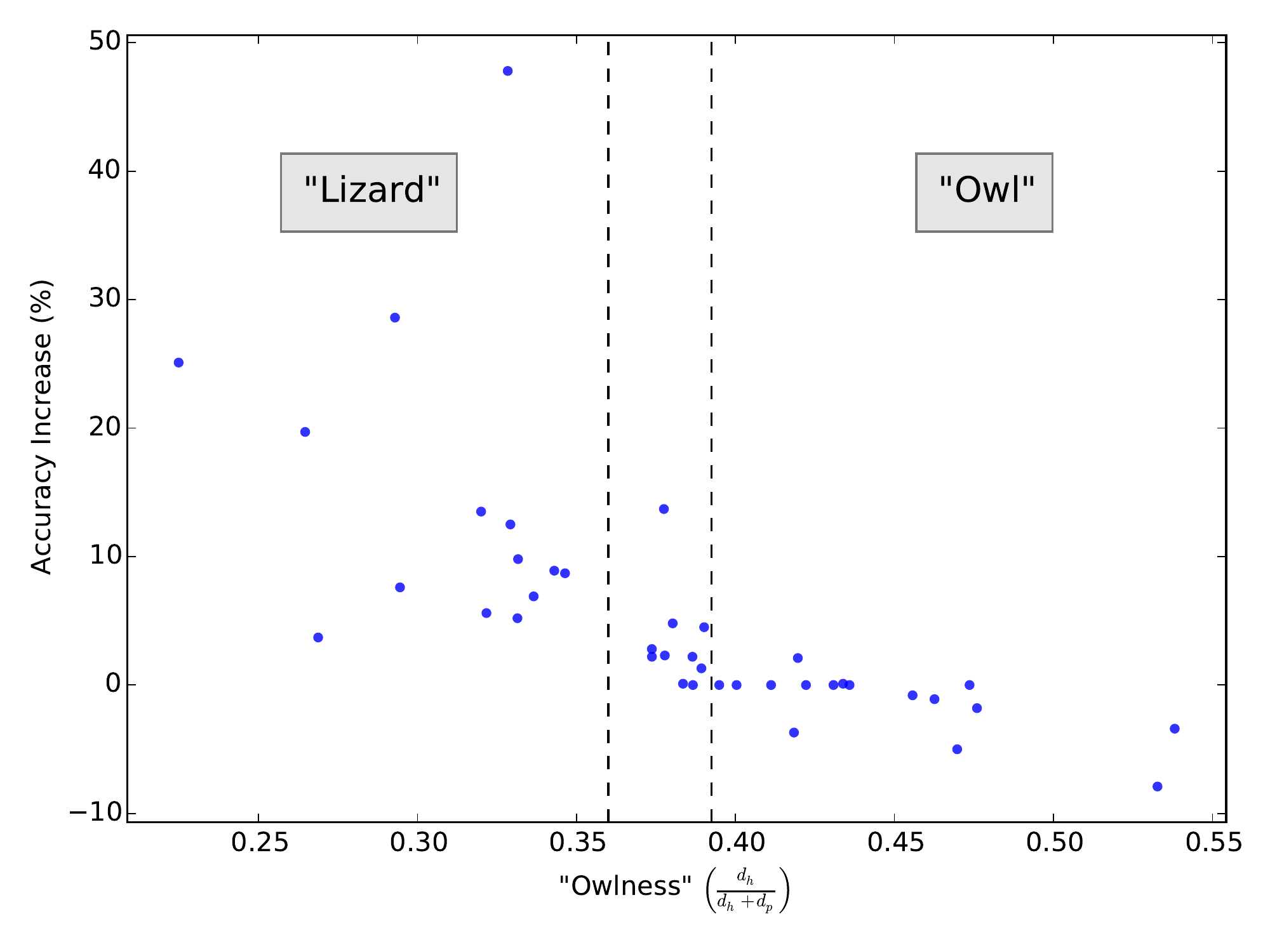}
    \caption{Increase in per-user accuracy versus the ``owlness'' metric which measures the fraction of the attention
      shift that is due to head movement versus eye movement. There are 40 points on this plot and each represents the
      average increase in accuracy achieved and the average measure of ``owlness'' for the user.}
    \label{fig:correlation}
  \end{subfigure}
  \hspace{0.2in}
  \begin{subfigure}[t]{0.45\textwidth}
    \includegraphics[height=2.2in]{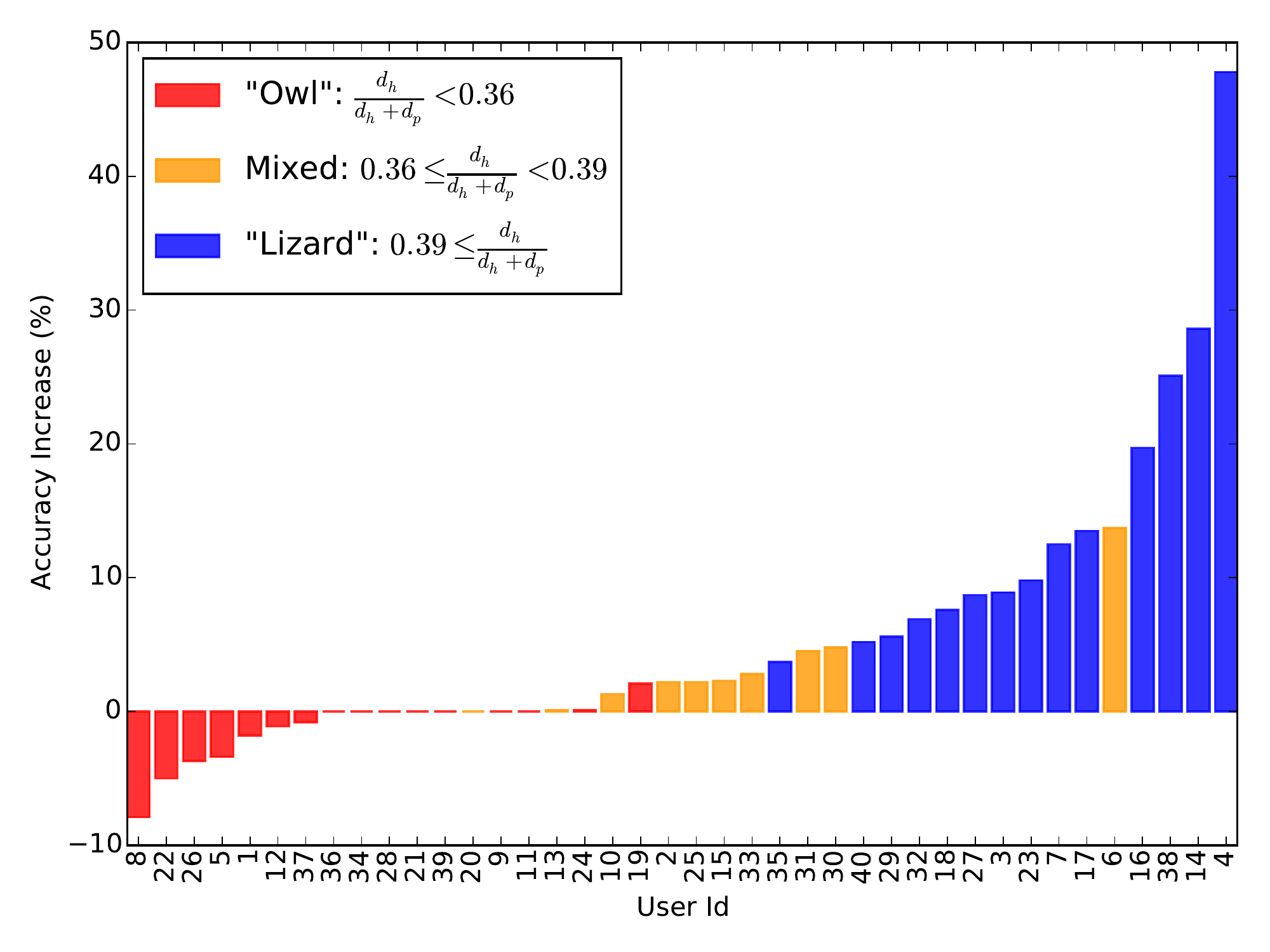}
    \caption{Increase in per-user accuracy partitioned by two threshold values in the ``owlness'' metric. The users with
    the high ``owlness'' measures, see zero or less improvement from adding in eye pose. The users with
    the low ``owlness'' measures, see positive improvement from adding in eye pose.}
    \label{fig:per-user-accuracy-increase}
  \end{subfigure}
  \caption{The ``owlness'' metric and its correlation with the increase in per-user accuracy from the first histogram
    plot to the second one in \figref{per-user-accuracy}.}
  \label{fig:bla}
\end{figure*}

Using the Pearson correlation coefficient as a guide, we programmatically explored over a million pairs of variables in
search of an answer to the question of what explains this difference in classification performance between users. Some
variables correlate with per-region accuracy but not overall. For example, average magnitude of off-center head movement
and pupil movement are good predictors of classification accuracy for the ``Right'' region with head pose alone and with
head and eye pose together, respectively. We were not able to find a measure of an individual that correlated highly
with overall classification accuracy, but there are a few variables that correlate with the increase in accuracy
achieved by adding in eye pose. The most interesting and intuitive one is a metric we refer to as ``owlness''. It is
defined as:

\begin{equation}\label{eq:owlness}
  M = \frac{d_h}{d_h + d_p}
\end{equation}

\noindent where $d_h$ and $d_p$ are the distance of the nose tip and right pupil, respectively, from their average
position in the background model. Due to the normalization of the features both distances are in the range
$[0, \sqrt{2}]$. An $M$ value of 0 means that a shift in gaze involves only the eyes (``lizard''). An $M$ value of 1
means that a shift in gaze involves only the head (``owl'').

\figref{correlation} shows the relationship between the ``owlness'' metric and the per-user increase in accuracy
achieved. The measure of ``owlness'' for each user is computed by averaging the result of \eqref{owlness} for each image
that passes the face detection and pupil detection steps in the pipeline. We partition users into three groups: ``owl'',
``lizard'', and ``mixed'' based on the value of $M$. \figref{per-user-accuracy-increase} shows how effectively these
partitions separate the users who gain classification accuracy from eye pose and those who do not. In this figure, the
``owls'' see no effect or a decrease in accuracy, while the ``lizards'' see a significant increase in accuracy.

\section{Conclusion}\label{sec:conclusion}

This paper investigates the contribution of head pose and eye pose to gaze classification accuracy for different gaze
strategies. We answer two questions: (1) how much does eye pose contribute and (2) how can the inter-user accuracy variation
be explained? For the former, we show that eye pose adds a 5.4\% increase in average accuracy (from 89.2\% to 94.6\%)
with an effective average rate of 1.3 decisions per second. For the latter, we propose an ``owlness'' metric that
decomposes gaze into head movement and eye movement and computes the relative magnitude of each. This metric
is used to explain the inter-person variation in impact of eye pose on gaze classification accuracy.

\section*{Acknowledgment}

Support for this work was provided by the Santos Family Foundation, the New England University Transportation Center,
and the Toyota Class Action Settlement Safety Research and Education Program. The views and conclusions being expressed
are those of the authors, and have not been sponsored, approved, or endorsed by Toyota or plaintiffs’ class
counsel. Data was drawn from studies supported by the Insurance Institute for Highway Safety (IIHS).

\bibliographystyle{IEEEtran}
\bibliography{eyes,lex-fridman,agelab}

\end{document}